\documentclass[pmlr]{jmlr}
\usepackage{longtable}% for long tables

\usepackage{booktabs}
\usepackage[load-configurations=version-1]{siunitx} % newer version
\theorembodyfont{\upshape}
\theoremheaderfont{\scshape}
\theorempostheader{:}
\theoremsep{\newline}

\title[multi-BLADE]{Exemplar Auditing for Multi-Label\\Biomedical Text Classification}

\author{\Name{Allen Schmaltz} \Email{aschmaltz@hsph.harvard.edu} 
       \addr Department of Epidemiology\\
       Harvard University\\
       Boston, MA, USA 
       \AND
       \Name{Andrew Beam} \Email{andrew\_beam@hms.harvard.edu} 
       \addr Department of Epidemiology\\
       Harvard University\\
       Boston, MA, USA } 

\usepackage{basecommon} 

\usepackage{multirow}
\usepackage{basecommon} 
\usepackage{array}
\newcolumntype{P}[1]{>{\centering\arraybackslash}p{#1}}
\newcolumntype{T}[1]{>{\raggedright\arraybackslash}p{#1}}
\begin{document}

\maketitle

\begin{abstract}
Many practical applications of AI in medicine consist of semi-supervised discovery: The investigator aims to identify features of interest at a resolution more fine-grained than that of the available human labels. This is often the scenario faced in healthcare applications as coarse, high-level labels (e.g., billing codes) are often the only sources that are readily available. These challenges are compounded for modalities such as text, where the feature space is very high-dimensional, and often contains considerable amounts of noise. 

In this work, we generalize a recently proposed zero-shot sequence labeling method, ``binary labeling via a convolutional decomposition'', to the case where the available document-level human labels are themselves relatively high-dimensional. The approach yields classification with ``introspection'', relating the fine-grained features of an inference-time prediction to their nearest neighbors from the training set, under the model. The approach is effective, yet parsimonious, as demonstrated on a well-studied MIMIC-III multi-label classification task of electronic health record data, and is useful as a tool for organizing the analysis of neural model predictions and high-dimensional datasets. Our proposed approach yields both a competitively effective classification model and an interrogation mechanism to aid healthcare workers in understanding the salient features that drive the model's predictions.
\end{abstract}

\section{Introduction}

A considerable amount of medical and scientific knowledge is encoded in unstructured, high-dimensional data. Much of this data is the result of human processes that produce human-mediated labels at a particular level of granularity, but then the investigator seeks analyses at a lower granularity of the data, for which there are no explicit ground-truth labels. With data modalities such as text, it is particularly challenging to uncover correlations across label granularities, given the complex dependencies among words and phrases.

Attention mechanisms, and related methods, provide a stochastic approach to relating document- and token-level scores in a test instance, but they are an incomplete solution to the broader challenge of interpreting black box neural models and data, since they do not provide a clear means of assessing the extent to which an example at test time is reflective of the data used to train the model.

Toward this end, the recent work of \citet{Schmaltz-2019-BLADE-1} proposed an approach for decomposing a document-level CNN binary classifier to produce token-level scores. This approach, ``binary labeling via a convolutional decomposition'' (\textsc{BLADE}), was shown to yield sharp token-level feature detections for a challenging zero-shot binary sequence labeling task. Importantly, this approach has the benefit that the token-level scores have a natural local, token-specific vector summarization derivable from the CNN filters. This enables an analysis method, exemplar auditing, for leveraging the CNN filters as representative keys of the strong class conditional feature detection of the binary \textsc{BLADE} model.

We extend the \textsc{BLADE} model and exemplar auditing to the multi-label classification setting and demonstrate its usage for medical text. We further examine the relative distances among exemplars in this context, proposing a novel, intuitive approach for analyzing these distances at an instance and label-specific level. We suggest that this analysis machinery is particularly applicable to medical settings, where the data is high-dimensional and noisy, and yet there is a need to have some level of verification of discovered patterns based on the available labeled training data.

\paragraph{Technical Significance}

We extend the previous work of \citet{Schmaltz-2019-BLADE-1} to the multi-label classification setting. Furthermore, we demonstrate an approach for applying exemplar auditing to the case when the input to the CNN is not a contextualized model, and where the relative rankings of the predicted multi-labels is important, combining losses for signal from both the local and global levels in a straightforward but effective manner. In this context, we demonstrate competitive classification results on a multi-label classification task of electronic health record data. Finally, we provide what is, to the best of our knowledge, a novel approach for analyzing and normalizing the distances from exemplar auditing, demonstrating that distances to nearest true positive, false negative, false positive, and true negative representative vectors from the training set provide useful signal in assessing a prediction. This is an intuitive and effective means of normalizing the distances for a given instance for a given label.

\paragraph{Clinical Relevance}

In order to put our methods in the context of previous work, we focus on annotating free text clinical notes with ICD-9 labels in patient discharge summaries. It is estimated that the United States spends in excess of many billions annually on unnecessary administrative costs such as from a complex billing infrastructure \citep{yong2010excess}, so improving and streamlining this process is of high value. This setting is emblematic of the more general setting where we have a large amount of high-dimensional data (such as text) about patients and labels at a high-granularity, but then we want to analyze the data at a lower-level of granularity. This could be for introspecting a prediction about a patient: We predict an outcome (or possible applicable diagnosis) for a patient with a model, but we want to examine the training set (for which we have known ground-truth labels) for similar patients to help inform our own clinical decision-making. It could also be used more generally for uncovering previously unknown patterns in large, high-dimensional datasets of patients or drugs, or for discovering label discrepancies in training datasets that may be used in high-stakes decision making.

We are proposing a rather different way of making sense of models and data than, say, interpreting coefficients on models, and/or examining p-values. We instead distill the relevant data into a small number of intuitive and information-dense values: localized features and associated predictions (and ground-truth, were available) from test and training, and the relative, normalized distances to the nearest true positive, false negative, false positive, and true negative representative examples from the training set. As a result, we can leverage the ability of the neural networks to find signals in large amounts of data, while retaining a straightforward means of ingesting and assessing those insights at a human level.

\section{Methods} 

We extend the convolutional decomposition proposed in \citet{Schmaltz-2019-BLADE-1}, which was originally evaluated on binary labeling settings, to the case where the labels are themselves high-dimensional. The approach utilizes a one-dimensional convolutional network \citep{Kim-2014-CNN}, which is then decomposed to produce scores at the token-level (i.e., the ``local'' level), even though token-level labels are not available during training. We start by describing the CNN, as used for classification at the document-level\footnote{In practice, the document can also be a single sentence. The key distinction is that the level of analysis of the base classifier is that at which human labels are available, and then the CNN is decomposed to produce scores at a lower level of granularity.} (i.e., the ``global'' level), before detailing the decomposition, associated loss functions, and exemplar auditing, the means of introspecting the predictions, for token-level analyses.

\paragraph{Multi-Label Classification}
%Each token (such as a word or punctuation symbol)
Each token $t_1,\ldots,t_n,\ldots,t_N$ in the document is represented by a $D$-dimensional vector, where $N$ is the length of the document, including padding symbols, as necessary. This $\reals^{D \times N}$ matrix vector is the input to the CNN. This mapping of a token to $D$-dimensional vectors can be, for example, via standard word embeddings \citep{PenningtonEtAl-2014-Glove,MikolovEtAl-2013-Word2vec}, a concatenation of standard word embeddings and contextualized embeddings \citep{DevlinEtAl-2018-BERT}, or in principle, a neural network over other input modalities pre-trained with a masked-language-model-style, or related, loss (over images, time-series data, etc.). In this work, we consider the first input modality.

The convolutional layer is applied to this $\reals^{D \times N}$ matrix, using a filter of width $K$, sliding across the $K$-sized ngrams of the input. The convolution results in a feature map $\boldh_m \in \reals^{N-K+1}$ for each of $M$ total filters. Note that each of the filters has a bias and $(D \cdot K)$ weights. 

We then compute
\begin{align*}
g_m &= \max \text{ReLU}(\boldh_m), 
\end{align*}
a ReLU non-linearity followed by a max-pool over the ngram dimension resulting in $\boldg \in \reals^M$. A final linear fully-connected layer, $\boldW \in \reals^{2C \times M}$, with a bias, $\boldb \in \reals^{2C}$, produces a vector of scores, $\boldo \in \reals^C$, for each of the $C$ class labels: 
\begin{align*}
\boldo &= \boldW_{1:C,1:M}\boldg + \boldb_{1:C}-\boldW_{C:2C,1:M}\boldg - \boldb_{C:2C}. 
\end{align*}

Typically such classifiers are trained for document classification (with similar effect) with a fully connected layer of dimension $\boldW \in \reals^{C \times M}$. Here, we have replaced $C$ with $2C$ (with a concomitant subtraction of the output) as a minor change to maintain the semantics of the binary decomposition (i.e., each label has a positive, or ``on'', state and a negative, or ``off'', state). Using this convention, the ``on'' weights of label $c \in C$ are in row $c$ of $\boldW$, and the ``off'' weights of label $c \in C$ are in row $2c$ of $\boldW$ (and analogously for the bias).

Here, we have a multi-label setting (i.e., each document can be assigned multiple labels) rather than a multi-class setting (i.e., with exclusive assignment of one label from a set of 2 or more labels); as such, we train with a sigmoid transform and a binary cross-entropy loss, as opposed to a softmax cross-entropy as used in multi-class settings:

\begin{align*}
L_c &= -Y_c\cdot \log \sigma(o_c) - (1-Y_c)\cdot \log (1-\sigma(o_c)), 
\end{align*}
where $Y_c \in \{0,1\}$ is the corresponding true class assignment for label $c \in C$ (at the document level). This loss is averaged over all $C$ classes, over the documents in the mini-batch.
 
 \paragraph{CNN Decomposition: \textsc{multi-BLADE}}
 
 We seek token-level scores \textit{for every label}, which we obtain by decomposing the final layer CNN. We use the notation
\begin{align*}
n_m &= \argmax \text{ReLU}(\boldh_m), 
\end{align*}
to identify the index into the feature map $\boldh_m$ that survived the max-pooling operation, which corresponds to the application of filter $m$ starting at index $n_m$ of the input (i.e., the set $\{n_m,\ldots,n_m+(K-1)\}$ contains all of the indices of the input covered by this particular application of the filter of width $K$). Note that the filter output is constant across labels, but each label is associated with a unique set of weights (and a bias) from the final fully-connected layer. We obtain a positive (``on'' state) contribution score $s^{c+}_n \in \reals$ for each input token $n$, for each label $c$, as follows:
\begin{align*}
s^{c+}_n &= (\Sigma_{m=1}^{M}W_{c,m}\cdot g_m\cdot\Sigma_{k=1}^K [n=n_m+(k-1)]) + b_c. 
\end{align*}
where we have used an Iverson bracket for the indicator function. The corresponding negative (``off'' state) contribution score $s^{c-}_n$ for token $n$ and label $c$ is analogous:
\begin{align*}
s^{c-}_n &= (\Sigma_{m=1}^{M}W_{2c,m}\cdot g_m\cdot\Sigma_{k=1}^K [n=n_m+(k-1)]) + b_{2c}, 
\end{align*}
\paragraph{Fine-Tuning: Multi-Label Min-Max + Global Normalization}

The token-level scores can then be used directly (perhaps with some lightweight tuning of the bias---i.e., the decision threshold---by an end-user\footnote{End-user tuning can also be useful for the fine-tuned models and is simpler and more perfunctory than it may sound: In practice, a clinician, or data annotator, is given access to the output in an interface with a single ``slider'', or other mechanism, to adjust a single real value to offset the learned bias, which has the effect of modulating precision and recall. Since the preferred balance between precision and recall varies across settings and end-users, such a mechanism is likely necessary in practice. We leave this for future HCI studies to investigate further.}). However, the decomposition affords flexibility in defining additional loss constraints, with which we can fine-tune the model, that can be useful in practice.

In documents associated with a label, we assume that at least one token in the document is associated with the label (i.e., the positive contribution score is greater than the negative contribution score for at least one token), but at least some tokens (and in fact, perhaps most, in practice) are not primarily associated with the label (in the sense that an end-user would not label the tokens with that class, but of course, there could be indirect dependence). Similarly, for documents not associated with a label, we assume all of the token-level positive contribution scores are less than or equal to the negative contribution scores for that label. We can encode this in the following min-max loss over labels\footnote{This is a generalized form of the two-class min-max binary cross-entropy loss of \citet{Schmaltz-2019-BLADE-1}, which was adapted from the two-class min-max squared loss of \citet{ReiAndSogaard-2018-ZeroShotSeq} over attention for grammatical error detection.}:   
\begin{align*}
L^c_{min} &= -\log (1-\sigma(s^{c+-}_{min})), 
\end{align*}
where $s^{c+-}_n = s^{c+}_n-s^{c-}_n$ is a combined token contribution and $s^{c+-}_{min}=\min(s^{c+-}_1,\ldots,s^{c+-}_n,\ldots,s^{c+-}_N)$ is the smallest combined token contribution in the document for label $c$; and
\begin{align*}
L^c_{max} &= -Y_c\cdot \log \sigma(s^{c+-}_{max}) - (1-Y_c)\cdot \log (1-\sigma(s^{c+-}_{max})), 
\end{align*}
where $s^{c+-}_{max}=\max(s^{c+-}_1,\ldots,s^{c+-}_n,\ldots,s^{c+-}_N)$ is the largest combined token contribution in the document for label $c$ and $Y_c \in \{0,1\}$ is the corresponding true class assignment for label $c \in C$.

Just fine-tuning using the aforementioned min-max loss can yield strong F-scores at the label level (derived from the max token contribution scores), but the contribution scores across labels (at the document level) may lack the normalization of the original fully-connected layer. In other words, the contribution scores for label $c$ may be reasonable at the token level, but comparing $\sigma(s^{c+-}_{max})$ with $\sigma(s^{(c+1)+-}_{max})$ may be a less reliable measure of the relative propensity for label $c$ vs. $(c+1)$ at the document level than just comparing $\sigma(o_c)$ with $\sigma(o_{c+1})$ resulting from training with the standard binary cross-entropy loss. This can be an issue if we aim to evaluate labels at the document level with retrieval-style ranking metrics---or in practice, aim to only present the end user with the subset of the top $k$ most likely labels for the document. One resolution to this issue is to simply ensemble the originally trained model and the min-max fine-tuned weights (as for example, using the former for label ranking and the latter for visualizing token-level scores), but we can alternatively modify the loss to incorporate the intuition of this ensemble approach, which has the benefit of generating a single, shared set of CNN filter weights for use with our analysis methods.

To enforce this global constraint, after training the base model and prior to fine-tuning, we instantiate a second linear fully-connected layer, $\boldW' \in \reals^{2C \times M}$, with a bias, $\boldb' \in \reals^{2C}$, with un-tied weights copied from $\boldW$ and $\boldb$, respectively. These two linear layers share the same convolutional filters (and input to the CNN), but the weights and biases are free to change separately. We then consider the following additional loss:
\begin{align*}
L^c_{combined} &= -Y_c\cdot \log \sigma(o'_c+s^{c+-}_{max}) - (1-Y_c)\cdot \log (1-\sigma(o'_c+s^{c+-}_{max})), 
\end{align*}
where $\boldo'$ is calculated in the same manner as $\boldo$ (in the base model), but with $\boldW'$ and $\boldb'$ instead of $\boldW$ and $\boldb$.

$L^c_{min}$, $L^c_{max}$, and $L^c_{combined}$ are then averaged over all classes over all documents in the mini-batch. In this way, local sparsity is enforced on the token-level scores (via $L^c_{min}$ and $L^c_{max}$), and global normalization across labels is maintained with the document-level maxpooling inherent in the calculation of $\boldo'$. Both the local and global constraints interact in $L^c_{combined}$.

\paragraph{Inference}

At inference, we assign label $c$ to the document if $\sigma(o'_c+s^{c+-}_{max}) > 0.5$.

\paragraph{Visualization of Token-Level Scores}

When visualizing the token-level score for token $n$ for label $c$ (i.e., the token-level label assignment) of a model fine-tuned just with the $L^c_{min}$ and $L^c_{max}$ losses, we find that $s^{c+-}_n > 0$ is a natural baseline decision threshold. When using $L^c_{min}$, $L^c_{max}$, and $L^c_{combined}$, we instead use the following:
\begin{align*}
o'_c + s^{c+-}_n > 0 \text{ if } s^{c+-}_n > 0 \text{ else False},
\end{align*}

which takes into account the addition of the global score.

\paragraph{Exemplar Auditing}

The previous work of \citet{Schmaltz-2019-BLADE-1} proposed exemplar auditing, an approach for leveraging the CNN filters as representative keys of the strong class conditional feature detection of the binary \textsc{BLADE} model, affording a means to introspect the training set (hereafter, ``database'') for a nearest neighbor to a relevant local feature in a test (hereafter, ``query'') prediction. This can be useful to audit the prediction, either for labeling additional data, or more generally for analyzing the data and model behavior. We further explore this idea in the context of multi-label classification.

Each token is associated with a vector that corresponds to the relevant filter applications from the convolution. In order to consider filters of arbitrary width, we associate a token with the average of all filter applications that covered the token (prior to the global maxpool operation). More specifically, with $M$ filters of width $K$, for each token $n$ we have a vector $\boldv^K_n\in \reals^M$:
\begin{align*}
\boldv^K_n = \frac{\sum{h_{1,n}+\ldots+h_{1,n+(K-1)}}}{K},  \\  
                  \ldots,
                  \frac{\sum{h_{m,n}+\ldots+h_{m,n+(K-1)}}}{K}, \\
                  \ldots,
                  \frac{\sum{h_{M,n}+\ldots+h_{M,n+(K-1)}}}{K},
\end{align*}
where we have averaged all components of each of the $M$ feature maps that resulted from an application over the token at index $n$. In the case of multiple filter widths, we concatenate all of the resulting vectors.

Since in our experiments the input to the CNN is not necessarily a contextualized embedding that has access to the full document, and since our inference scoring takes into account the maxpool vector (via $o'_c$), we also consider the document-level maxpool vector (with ReLU\footnote{The motivation for not applying ReLU and not censoring the negative components in $\boldv^K_n$ is that there is potentially informative signal in the negative values for distinguishing exemplars at the n-gram level. In the case of $\boldv^{maxpool}$, we use the ReLU for consistency with the maxpooling of training (and in any case, padding tokens and n-grams impose a de-facto ReLU, unless masked, since they are zero by default).}) $\boldv^{maxpool} \in \reals^M$:
\begin{align*}
\boldv^{maxpool} = \max \text{ReLU}(h_{1}), \ldots, \max \text{ReLU}(h_{m}), \ldots, \max \text{ReLU}(h_{M}),
\end{align*}
which is constant for all tokens in the document. The full vector $\boldv_n$ for the token at index $n$ is then the concatenation of the applicable token-specific components of the feature maps and the document-level maxpool components:
\begin{align*}
\boldv_n = \boldv^K_n, \boldv^{maxpool}.
\end{align*}
We subsequently use $\boldv$ to refer to exemplar vectors from the database, and we use $\boldq$ to refer to such vectors from a query. As an important distinction, $\boldv$ has access to the ground truth labels from training, whereas $\boldq$ does not. Since operating over all exemplar vectors from every token in the database can be computationally expensive in both time and space for large numbers of long documents, we make the restriction that we only store one exemplar vector (with the max token-level contribution score) for each predicted or gold label for each document in the database. In other words, for a given class $c$ for a given document, we only store the exemplar vector corresponding to $s^{c+-}_{max}$ when $\sigma(o'_c+s^{c+-}_{max}) > 0.5$ and/or $Y_c \in \{1\}$. The number of predicted and gold labels per document is typically considerably less than both the total number of classes and the total number of tokens (and the same exemplar vector can be associated with multiple labels, but not vice-versa), so this restriction dramatically decreases the size of the database.

When classifying new documents at test time, for any class with a positive prediction, $\sigma(o'_c+s^{c+-}_{max}) > 0.5$, we can associate the query token $j$  with the exemplar vector at index $i$ from the database, $\boldv_i$, (and corresponding document) that minimizes the Euclidean distance with that of the query token's vector $\boldq_j$:
\begin{align*}
\argmin_{\boldv_i}\| \boldq_j-\boldv_i \|_2.
\end{align*}
Previous work demonstrated that the exemplars could be effectively utilized at inference by combining the query and database predictions via a conjunctive decision rule to increase the precision of the predictions. For reference (and as a means of organizing one's analysis), we also consider a soft combination between the query prediction and the prediction of the matched database exemplar modulated by relative distances. In practice when classifying new documents at test time, for the predicted class $c$ for each query token's vector, $\boldq_j$, we retrieve up to 4 distinct database vectors $\boldv_{i^1}, \boldv_{i^2}, \boldv_{i^3}, \boldv_{i^4}$, each of which corresponds to a unique document in the database:

\begin{enumerate}
\item The vector $\boldv_{i^1}$ minimizes the Euclidean distance with that of the query token's vector $\boldq_j$ with the restriction that $\boldv_{i^1}$ is associated with both a positive model prediction for class $c$ and a corresponding positive ground truth label for class $c$ (i.e., this is a true positive in the training set).
\item The vector $\boldv_{i^2}$ minimizes the Euclidean distance with that of the query token's vector $\boldq_j$ with the restriction that $\boldv_{i^2}$ is associated with a positive ground truth label for class $c$ but a negative model prediction for class $c$ (i.e., this is a false negative in the training set).
\item The vector $\boldv_{i^3}$ minimizes the Euclidean distance with that of the query token's vector $\boldq_j$ with the restriction that $\boldv_{i^3}$ is associated with a positive model prediction for class $c$, but a negative ground truth label for class $c$ (i.e., this is a false positive in the training set).
\item The vector $\boldv_{i^4}$ minimizes the Euclidean distance with that of the query token's vector $\boldq_j$ with the restriction that $\boldv_{i^4}$ is associated with both a negative model prediction for class $c$ and a negative ground truth label for class $c$.\footnote{Given our database restrictions to reduce computational costs, in practice this case retrieves a token for which the document is not associated with class $c$ (either as a prediction or ground-truth label), but the token is associated with at least one other class label (either as a prediction or ground-truth label) as the max token-level contribution score.}
\end{enumerate}

Of these four vectors\footnote{It is possible for one or more of these vectors to not exist in the database (e.g., $\boldv_{i^1}$ will not exist if the model never correctly predicted that label in training), in which case we simply assign a very large default distance.}, we use the notation $\boldv_{i^*}$ to identify that which is the overall minimal distance to the query:
\begin{align*}
\boldv_{i^*}= \argmin_{\boldv_i}\| \boldq_j-\boldv_i \|_2.
\end{align*}
As one means of analyzing whether these pieces of information provide signals in the expected directions, we also provide results where at inference, if the query $\sigma(o'_c+s^{c+-}_{max}) > 0.5$, we assign label $c$ to the document if
\begin{align*}
 \sigma \left( o'_c+s^{c+-}_{max} +
database_{score} \cdot \frac{\exp(-\| \boldq_j-\boldv_{i^*} \|_2)}{\sum^4_{z=1} \exp(-\| \boldq_j-\boldv_{i^z} \|_2)} \right) > 0.5,
\end{align*}
where (with a slight overloading of notation between scores from the query and database), $database_{score}$ is the model score $(o'_c+s^{c+-}_{max})$ associated with $\boldv_{i^*}$ from training (i.e., from running the model on that training document).

The resulting score could be used as a blind, automatic substitute for the original model score (and we provide results to this effect below for context), but that is not the intended use case. Rather, this machinery is a way of organizing a human end-user's evaluation of a model prediction (and the data), as analyzed further below.

Note that the softmax over the negative distances has the effect of down-weighting the impact of the score from the database when the other exemplars have relatively similar distances. In a high-stakes scenario, an end-user could instead impose a hard rejection of a label if, for example, the exemplar $\boldv_{i^*}$ was not $\boldv_{i^1}$, or use that as reference context for re-labeling the data. 

\section{Experiments} 

Our approach can be generally applied to any multi-label text classification task. In the interest of comparing to previous work in the medical domain with data that is available for replication, we focus on the clinical text from the previous work of \cite{mullenbach-etal-2018-explainable}.

\paragraph{Data and Task}

The Medical Information Mart for Intensive Care (MIMIC-III) dataset version 1.4 \citep{Johnson-etal-2016-MIMIC-III-main-publication,Pollard-etal-2016-MIMIC-III-data,Physionet-ref-for-MIMIC-3} is a large-scale dataset of de-identified patient data derived from admissions to a Boston-area hospital. The dataset is available to researchers under a data use agreement. We focus on the text of the patient discharge notes, which are labeled with International Classification of Diseases (ICD-9) codes. These codes are primarily for billing and administrative purposes, but serve as a useful testing ground for high-dimensional multi-label classification in the medical setting given the availability of data and previous works for comparison. We hypothesize that many of the challenges involved in this reasonably well-defined, replicable setting will be present in other medical classification scenarios, and that it is thus a reasonable testing grounds on which to focus.

More specifically, the task is to assign one or more ICD-9 codes to each discharge summary. We follow the publicly available MIMIC-III preprocessing and setup of \cite{mullenbach-etal-2018-explainable}\footnote{We use the preprocessing code available at \url{https://github.com/jamesmullenbach/caml-mimic}.}, which lowercases and truncates the documents to a maximum length of 2500, removing any tokens that lack at least one alphabetic character. Low-frequency tokens (those occurring in less than three training documents) are replaced with a placeholder symbol. We use a comparable vocabulary size of the 50k most common tokens.

We follow past work and provide results on two subsets of the data. In the first, we restrict the data to the top 50 most common labels, and only consider the 8066, 1573, and 1729 discharge summaries (hereafter, documents) associated with those labels in each of the train, development, and test sets, respectively. We also consider the full set of  8921 labels seen in the documents, which includes 47723, 1631, and 3372 documents in each of the train, development, and test sets, respectively. In this setting, there are 73 labels in the development set and 172 labels in the test set that are never seen in training (reflective of the larger universe of available ICD-9 codes). We follow past work in assigning these labels as missed predictions for the model at inference time.

The task is challenging for at least three reasons:
\begin{enumerate}
\item The label space is high-dimensional, with documents assigned a variable number of labels. For reference, in the test set in the top 50 labels subset, there are on average 6 labels assigned to every document, ranging from a minimum of 1 label to a maximum of 20 labels. In the test set for the full set, there are on average 18 labels assigned to every document, ranging from 1 to 65 labels.
\item The data is noisy, consisting of many incomplete, grammatically incorrect sentences with various abbreviations and medical-domain-specific language. Headings and other structures from the EHR are flattened into the text. As a result of the aforementioned, the text differs considerably from standard text used to pre-train typical NLP models. Additionally, while the number of documents may seem modest, in fact there is a considerable amount of text owing to the long length of the documents (as opposed to the ``documents'' consisting of single sentences). 
\item Owing to points (1) and (2) above, the task is also non-trivial for humans, introducing potential ambiguity and noise into the ground-truth labels.
\end{enumerate}

We assess our approaches using the metrics of the previous works analyzing this dataset, where for consistency we have used the same evaluation scripts of \cite{mullenbach-etal-2018-explainable}. This includes the micro-averaged and macro-averaged $F_1$ and the area under the ROC curve (AUC). Following the previous work, we focus on the retrieval metric, precision @ $z$ ($P@z$), as our primary metric. In this context, this metric is the average number of $z$ highest-scoring labels out of $z$ that are true labels in the ground truth data.\footnote{In the implementation of this metric in \cite{mullenbach-etal-2018-explainable}, the denominator was calculated as a constant $z$ across documents, which means that the gold labels will not yield a $P@z$ value of 1 against the ground truth, since some documents have less than $z$ true labels. In practice, we found that adjusting the denominator to the real number of true labels (when less than $z$) did not change the direction of any of the results in relative terms, so we stick to the previous formulation for consistency purposes.} This metric is chosen under the assumption that the real-world use case for such models is as an annotation support tool, emphasizing precision over recall, with the additional consideration that the relative ranking of predicted labels is important. In other words, we aim for a system that predicts labels that are relevant and true, with a ranking to allow an end-user to review the top few predicted labels. For the top 50 subset, we chose model parameters and perform tuning on the held-out development set using $P@5$, and similarly, $P@8$ for the full set.

In analyzing our proposed approaches, we aim for an input modality to the \textsc{multi-BLADE} layer that yields levels of effectiveness that are at least competitive with previous works. As we show below, the input word embeddings of previous work serve this purpose. We then use that as the substrate upon which to consider exemplar auditing. Note that the input (i.e., the underlying model of the bottom layers of the network) to the \textsc{multi-BLADE} layer is orthogonal to exemplar auditing in so far that we would assume that if there existed a significantly stronger model, it could be co-opted by incorporating the frozen version as input.

We turn now to the details of the models used in the experiments.\footnote{In the published version, we will include a link to the replication code.} 

\paragraph{CNN Model}

As our base model, we use a CNN with 100 filters of width 1 and 1000 filters each for widths of 3, 4, and 5. We train with Adadelta \citep{Zeiler-2012-Adadelta}, with dropout of 0.5 on the input to the final fully-connect layer, choosing the epoch with the highest $P@5$ score on the held out development set. We use the label \textsc{CNN$_{1345}$} to refer to this model. With the full set of labels, we use a similar model, but increase the model capacity to 200 filters of width 1 and 2000 filters each for widths of 3, 4, and 5. We found that training on the full set was very sensitive to model parameters. Based on results on the development set, we train with Adam \citep{Kingma-Ba-2014-Adam} using a small learning rate of $0.0001$ and dropout of 0.6. Additionally, we train with a schedule such that the model only considers the top 1000 most frequent labels for the first 30 epochs before transitioning to training with the full label set. As in previous work, we choose the epoch with the highest $P@8$ score on the development set. We use the label \textsc{CNN$_{1345}$+full} to refer to this model. With all models, we use pre-trained, 100 dimensional Word2Vec embeddings \citep{MikolovEtAl-2013-Word2vec} over the documents as in the work of \cite{mullenbach-etal-2018-explainable}.

\paragraph{CNN Fine-Tuning}

We fine-tune the base models using the $L^c_{min}$, $L^c_{max}$, and $L^c_{combined}$ losses, for which we use the labels \textsc{CNN$_{1345}$+mmc} and \textsc{CNN$_{1345}$+full+mmc}. In the case of \textsc{CNN$_{1345}$+full+mmc}, based on results on the development set, we only calculate token-level scores (and assign non-zero loss scores) for the top 1000 labels predicted by $\boldo'$ for each training instance in the mini-batch.

\paragraph{Exemplar Auditing}

The exemplar auditing machinery is primarily intended as a per-document level analysis tool for a human end-user. To assess the quality of the signals presented to the end-users, we provide empirical results using the same aggregated metrics as the core models. We label experiments using the aforementioned soft combination of query and database scores (and distances) with the suffix \textsc{+ExA}. In further analyses, we also consider a decision rule in which we only admit a prediction for a label if the retrieved $\boldv_{i^*}$ exemplar vector is $\boldv_{i^1}$, for which we use the label \textsc{+ExADR}. Finally, to provide a further evaluation of the similarity between the query and database vectors, we show results in which for a given model prediction, we substitute the score from the model on the query (i.e., the test set) with the score associated with $\boldv_{i^*}$ (i.e., the score from the training set). We label these experiments with \textsc{+onlyDB}. Note that the exemplars for the \textsc{CNN$_{1345}$+mmc} model are $\reals^{6200}$ vectors, and for \textsc{CNN$_{1345}$+full+mmc}, $\reals^{12400}$ vectors.

\paragraph{Previous Models}

The previous work of \cite{mullenbach-etal-2018-explainable} considers replacing the standard maxpooling of the base CNN with a per-label attention mechanism, which in effect is a learned weighted average over the filters, specific to each label. This model is referred to as Convolutional Attention for Multi-Label classification (\textsc{CAML}). A variant (\textsc{DR-CAML}) is also considered which regularizes the predictions using embeddings of the labels. Both \textsc{CAML} and the decomposition examined here can be used to generate token-level scores; however, the manner of doing so is rather different. Whereas \textsc{CAML} utilizes a softmax attention mechanism, \textsc{multi-BLADE} is a method of leveraging the maxpooling behavior of the base classifier, and can also be used to derive token-level scores without additional parameters to the base model (including if fine-tuned with only the min-max loss). Finally, we also consider the \textsc{LEAM} model of \cite{Wang-etal-2018-ACL-Label-Embeddings}, which learns a joint embedding attention between both the document text and the label text.\footnote{As suggested in passing above, we could also use frozen versions of these alternative models as input to the \textsc{multi-BLADE} layer. However, as we show below, using standard word embeddings as input already yields at least competitive results on the primary metrics of interest, so we do not pursue this avenue further in this work. In preliminary experiments, we found that using the frozen contextualized embeddings of \cite{DevlinEtAl-2018-BERT} led to a significant degradation in effectiveness, almost certainly owing to the large divergence between the domains on which these models were trained and the non-standard language of the discharge summaries. We leave retraining the contextualized embeddings on this domain of text to future work.} To our knowledge, the results in these works constitute the current baselines on this particular MIMIC-III task.

\section{Results}

In the analysis of the experimental results, we demonstrate the following two high-level points: 
\begin{enumerate}
\item We show that the proposed model and losses are at least competitive with previously proposed approaches on the main metrics on these datasets.
\item We show empirical evidence that the signals provided by exemplar auditing (as would be presented to an end-user at a per-document level) behave as expected.
\end{enumerate}

\begin{table}
\centering
%\footnotesize
\begin{tabular}{lcccc>{\bfseries}c}
\toprule
 & \multicolumn{2}{c}{AUC} & \multicolumn{2}{c}{$F_1$} & \\
Model & Macro & Micro & Macro & Micro & $P@5$\\
\midrule
\textsc{LEAM} & 0.881 & 0.912 & 0.540 & 0.619 & 0.612 \\
\midrule
\textsc{CAML} & 0.875 & 0.909 & 0.532 & 0.614 & 0.609 \\
\textsc{DR-CAML} & 0.884 & 0.916 & 0.576 & 0.633 & 0.618 \\
\midrule
\textsc{CNN$_{1345}$} & 0.910 & 0.935 & 0.586 & 0.655 & 0.652 \\
\midrule
\textsc{CNN$_{1345}$+mmc} &  0.913 & 0.937 & 0.598 & 0.663 & 0.654 \\
\midrule
\textsc{CNN$_{1345}$+mmc+ExA} & 0.913 & 0.937 & 0.591 & 0.658 & 0.652 \\
\bottomrule
\end{tabular}
\caption[Main results top 50]{MIMIC-III test set results on the top 50 labels. The \textsc{CAML} and \textsc{DR-CAML} model results are as reported in \citet{mullenbach-etal-2018-explainable}. $P@5$ (bolded column) is the metric used to tune the models on the development set.} 
\label{table:test-results-top50}
\end{table} 

\begin{table}
\centering
%\footnotesize
\begin{tabular}{lcccc>{\bfseries}cc}
\toprule
 & \multicolumn{2}{c}{AUC} & \multicolumn{2}{c}{$F_1$} & & \\
Model & Macro & Micro & Macro & Micro & $P@8$ & $P@15$ \\
\midrule
\textsc{CAML} & 0.895 & 0.986 & 0.088 & 0.539 & 0.709 & 0.561 \\
\textsc{DR-CAML} & 0.897 & 0.985 & 0.086 & 0.529 & 0.690 & 0.548 \\
\midrule
\textsc{CNN$_{1345}$+full} & 0.806 & 0.972 & 0.035 & 0.447 & 0.691 & 0.531 \\
\midrule
\textsc{CNN$_{1345}$+full+mmc} & 0.790 & 0.969 & 0.040 & 0.467 & 0.697 & 0.538 \\
\midrule
\textsc{CNN$_{1345}$+full+mmc+ExA} & 0.790 & 0.969 & 0.034 & 0.454 & 0.696 & 0.537 \\
\bottomrule
\end{tabular}
\caption[Main results all labels]{MIMIC-III test set results on all 8921 labels. The \textsc{CAML} and \textsc{DR-CAML} model results are as reported in \citet{mullenbach-etal-2018-explainable}. $P@8$ (bolded column) is the metric used to tune the models on the development set.} 
\label{table:test-results-all-labels}
\end{table} 

With regard to (1), Table~\ref{table:test-results-top50} displays the main results for the top 50 labels subset of the data. Of note is that on this subset, the benefits of the previously proposed attention mechanisms (\textsc{CAML} and \textsc{DR-CAML}) and label embedding approaches (\textsc{LEAM}) are within parameter variation of the base CNN model. 

In the top 50 labels set, the addition of the min-max loss (\textsc{+mmc}) does not lead to a real difference in the primary metric of interest ($P@5$). However, the key advantage of using this loss is that it does not degrade these document-level scores, but it does encourage sparsity in the scores at the token-level, which can be helpful when visualizing the output. This can be useful in practice when the approach is used as an annotation support tool.

The analogous results on the full set are shown in Table~\ref{table:test-results-all-labels}. In this case, the training of \textsc{CAML} appears to have found a particularly effective setting in the parameter space. In general, we found training on this full set to be very sensitive to minor changes in learning parameters (optimizer, learning rates, dropout probability, etc.), perhaps owing to the very long tail of infrequently occurring labels. Nonetheless, we find that the effectiveness of \textsc{CNN$_{1345}$+full+mmc} in terms of $P@8$ (the metric we tuned against on the development set) to be between that of \textsc{DR-CAML} and \textsc{CAML}, and to be competitive for all practical purposes. (Along these lines, note, too, that the relative effectiveness of \textsc{DR-CAML} and \textsc{CAML} flips across the top 50 subset and the full label set.)

\begin{table}
\centering
%\footnotesize
\begin{tabular}{l>{\bfseries}ccc>{\bfseries}ccc}
\toprule
 & \multicolumn{3}{c}{Macro} & \multicolumn{3}{c}{Micro} \\
Model & Precision & Recall & $F_1$ & Precision & Recall & $F_1$ \\
\midrule
\textsc{CNN$_{1345}$+mmc} & 0.704 & 0.520 & 0.598  & 0.765 & 0.586 & 0.663  \\
\midrule
\textsc{CNN$_{1345}$+mmc+ExA} & 0.705 & 0.508 & 0.591 & 0.769 & 0.575 & 0.658 \\
\midrule
\textsc{CNN$_{1345}$+mmc+onlyDB} & 0.715 & 0.480 & 0.574 & 0.777 & 0.548 & 0.643 \\
\midrule
\textsc{CNN$_{1345}$+mmc+ExADR} & 0.712 & 0.432 & 0.538 & 0.784 & 0.501 & 0.611 \\
\bottomrule
\end{tabular}
\caption[Additional ExA metrics for top 50 labels]{MIMIC-III test set results on the top 50 labels with a breakdown of precision and recall with and without the various exemplar auditing decision rules. The precision columns are highlighted for discussion in the main text.} 
\label{table:additional-ExA-analyis-top50-labels}
\end{table}

\begin{table}
\centering
%\footnotesize
\begin{tabular}{l>{\bfseries}ccc>{\bfseries}ccc}
\toprule
 & \multicolumn{3}{c}{Macro} & \multicolumn{3}{c}{Micro} \\
Model & Precision & Recall & $F_1$ & Precision & Recall & $F_1$ \\
\midrule
\textsc{CNN$_{1345}$+full+mmc} & 0.062 & 0.029 & 0.040 & 0.727 & 0.343 & 0.467  \\
\midrule
\textsc{CNN$_{1345}$+full+mmc+ExA} & 0.057 & 0.025 & 0.034 & 0.760 & 0.324 & 0.454 \\
\midrule
\textsc{CNN$_{1345}$+full+mmc+onlyDB} & 0.055 & 0.022 & 0.031 & 0.777 & 0.294 & 0.426 \\
\midrule
\textsc{CNN$_{1345}$+full+mmc+ExADR} & 0.055 & 0.020 & 0.029 & 0.785 & 0.263 & 0.395 \\
\bottomrule
\end{tabular}
\caption[Additional ExA metrics for all labels]{MIMIC-III test set results on all 8921 labels with a breakdown of precision and recall with and without the various exemplar auditing decision rules. The precision columns are highlighted for discussion in the main text.} 
\label{table:additional-ExA-analyis-all-labels}
\end{table}

\begin{table}
\centering
%\footnotesize
\begin{tabular}{lcccc}
\toprule
 & \multicolumn{4}{c}{Softmax Threshold for $\boldv_{i^1}$}  \\
Model & 0.0 & 0.2 & 0.4 & 0.6 \\
\midrule
\textsc{CNN$_{1345}$+mmc+ExADR+t} & 0.784/0.501 & 0.784/0.501 & 0.871/0.208 &  0.984/0.018 \\
\midrule
\textsc{CNN$_{1345}$+full+mmc+ExADR+t} & 0.785/0.263 & 0.785/0.263 & 0.815/0.203 &  0.885/0.071  \\
\bottomrule
\end{tabular}
\caption[Additional ExADR thresholds]{Micro Precision/Recall on the MIMIC-III test set on the top 50 labels subsets and all 8921 labels, only admitting a label prediction based on \textsc{ExADR} and if the corresponding softmax distance probability is greater than the specified threshold.} 
\label{table:additional-distance-analysis-all-labels}
\end{table} 

With regard to analysis point (2) above, we see in Tables~\ref{table:test-results-top50} and ~\ref{table:test-results-all-labels} that the combination of the query scores and the database scores (with \textsc{+ExA}) does not significantly change the $P@5$ and $P@8$ scores. This provides evidence that the query scores and the exemplar scores (weighted by relative distances) tend to be in the same direction. We examine this further in Table~\ref{table:additional-ExA-analyis-top50-labels} for the top 50 labels subset and in Table~\ref{table:additional-ExA-analyis-all-labels} for the full label set where we break down the precision and recall values used to calculate the $F_1$ scores. We see that the precision of the predictions is generally retained when combining the query and database scores, and in fact, the Micro precision rises by around 3 points for \textsc{CNN$_{1345}$+full+mmc+ExA} relative to \textsc{CNN$_{1345}$+full+mmc}.

Interestingly, if we throw away the model prediction of the query and replace it with the database prediction associated with the exemplar, the $F_1$ scores only suffer a modest decline, and it is the result of a decline in recall but in fact is accompanied by a rise in precision, as we see in Tables~\ref{table:test-results-top50} and ~\ref{table:test-results-all-labels} for \textsc{+onlyDB}. Note that this is without constraining or censoring the choice of $\boldv_{i^*}$, and so the relative stability of this change is reflective of most exemplar vectors being associated with predictions in the same direction as the query. We also see that the hard decision rule of \textsc{+ExADR} tends to push up Micro precision. Most selected database exemplars are $\boldv_{i^1}$ vectors, which is why we see only a modest (and not catastrophic) decline in recall with the \textsc{+ExADR} decision rule. 

Exceptions to some of the above patterns are with the Macro metrics, which have the effect of heavily weighting (in relative terms) low-frequency labels.\footnote{As \cite{mullenbach-etal-2018-explainable} note, ``A hypothetical system that performs perfectly on the 500 most common labels, and ignores all others, would achieve a Macro $F_1$ of 0.052 and a Micro $F_1$ of 0.84.''} As with previous work, the values are sufficiently low in the full set (resulting from relatively rare correct predictions on the long tail of labels that occur infrequently in training) that the observed differences may not be meaningfully different in practice, and are thus difficult to draw conclusions from, beyond concluding that none of these models are particularly effective on rare labels, at least in the aggregate.    

It is also useful to have an empirical sense of whether the relative distances behave as expected. In particular, the relative distance associated with the $\boldv_{i^1}$ vectors should contain information regarding the reliability of the prediction. If the $\boldv_{i^1}$ vector is close to the query vector $\boldq_j$, and at the same time, $\boldq_j$ is comparatively far from each of $\boldv_{i^2}$, $\boldv_{i^3}$, and $\boldv_{i^4}$, we would expect the query prediction to be more likely to be right than if the distance to the $\boldv_{i^1}$ vector is farther in relative terms to the distances to $\boldv_{i^2}$, $\boldv_{i^3}$, and $\boldv_{i^4}$. A clean way of evaluating this is to simply vary a threshold on the normalized softmax distance for the $\boldv_{i^1}$ vectors. We show results in Table~\ref{table:additional-distance-analysis-all-labels} in which we only admit a label prediction if $\boldv_{i^*}$ is $\boldv_{i^1}$ \textit{and} the normalized softmax probability is greater than a given threshold. We label these results with \textsc{+ExADR+t}. (Recall that the softmax probability is derived from a negative distance, so a higher probability implies closer similarity.) Here we show a relatively coarse grid search, but the pattern is clear: As the query and $\boldv_{i^1}$ increase in similarity in relative terms to the distances to the false negatives, false positive, and true negative vectors, the Micro precision of the predictions rise.

Appendix A contains output from three random sentences from the test set for the \textsc{CNN$_{1345}$+full+mmc}, along with the exemplar vectors. Given the noisy nature of the data, the long documents, and high-dimensional label set, it is at times striking how sharp the feature detections are. Note that although we are not using contextualized embeddings as input to the CNN, as a result of the averaging over the filters to construct exemplar vectors, which means that a token with an application of a filter width of 5 sees filter applications over a total of 9 tokens, and the concatenation of the global maxpool vector, each vector $\boldq_j$ and $\boldv_i$ has a relatively expansive view of the document.

The above results encapsulate the crux of the approach: We can focus on representative local features (which are human interpretable, or at least human manageable as a means of pivots for organizing one's analysis) and exploit relative distances between summary vectors of true positive, false negative, false positive, and true negative elements of the training set in order to aid in the analysis of unidentifiable neural models and associated high-dimensional data. It is a surprisingly parsimonious, yet powerful idea that we expect will have a number of real-world applications.

\section{Limitations}

The primary limitation of the approach is that---relative to performing a standard forward pass with a CNN classifier---it can be relatively computationally expensive to search for the exemplars over a large database for many queries. However, we found in practice that the calculation of the distances remains practical provided the Euclidean distances are calculated on a GPU, noting that the exemplar search itself is embarrassingly parallel, which allows for straightforward splitting across multiples GPUs.

It is also important to reiterate (which should be clear from above) that utilizing exemplar auditing does not automagically make a decent classifier a significantly better classifier. As we show above, the various pieces of information can be used (if so desired) as a means of constraining predictions to boost precision (along the precision-recall curve), but we would not typically expect huge improvement swings in overall model prediction effectiveness in doing so over data similar to that seen in training (which in that way, would not be faithful to the underlying model, in any case), with the notable possible exception of the case of database exemplars over data not seen in explicit training. Rather, the approach is a means of providing a human with the key pieces of (likely applicable) information, among large amounts of possible information, to assess a model decision and its associated data. With that information, a human user can more effectively use the model as an assisting tool in decision-making.

\section{Related Work} 

In NLP, many surface-level interpretation methods have been proposed, often based on attention mechanisms. \cite{BelinkovAndGlass-2019-NLP-Analysis-Methods} provides a recent overview. Here, we premise our approach for relating document-level scores to token-level scores on the previous work of \citet{Schmaltz-2019-BLADE-1}, which demonstrated that for binary zero-shot grammatical error detection (a sequence labeling task for which ground-truth token-level labels are available), a decomposition of a CNN was at least competitive with previously proposed attention-based approaches. We extend the approach to the multi-label setting. High-dimensional multi-label classification opens a number of possibilities for adjacent tasks; in future work, we plan to explore the utility of this approach in regression settings via discretizing real-valued output.

The exemplar auditing concept of relating a fine-grained feature of a test instance back to a feature in training and utilizing relative distances to analyze a model and its data is a rather different notion of model interpretation than is typically considered in the attention-mechanism literature in NLP, and we think it is an important avenue for further work. This notion of relative distances bears some resemblance to---but is largely orthogonal to---the large literature of bayesian and frequentist approaches for calculating decision bounds. \cite{Card-etal-2019-ConformalNLP} propose a conformal-based method to describe a model prediction in terms of a weighted sum of training instances, where the measure of non-conformity is a distance between the final hidden state of a neural classifier (prior to the softmax). As an important distinction, their proposed approach relates predictions at the instance level (e.g., at the document level), whereas the machinery presented here provides a means of dissecting model predictions at the fine-grained feature level, which is critical for domains such as text, particularly when the documents are very long. Either the distance to the exemplars, or in fact, the softmax distribution over exemplar distances for an instance, itself, could be used as part of a non-conformity score in a conformal framework, which we leave to future work.

The prototypical networks of \cite{SnellEtAl-NIPS2017-PrototypicalNets} can be used for zero-shot and one-shot classification by assigning an instance to a cluster based on a softmax over distances to vectors representing the classes, which are means over the instances of the classes. With exemplar auditing combined with \textsc{[multi-]BLADE}, we instead retain granularity over the fine-grained features of the label classes (and document classifier) as represented by the exemplar vectors, which is a key difference. Additionally, our approach produces a softmax distribution over distances to the nearest true positive, false negative, false positive, and true negative representatives of \textit{a feature for a particular document for a particular label}, which is, to our knowledge a new approach that has not been previously explored. As we show above, these relative distances provide informative signal as to the reliability of the prediction.

\section{Conclusion}
 
We have examined an approach for organizing the analysis of a multi-label classifier and its associated data, using a CNN as the final layer of a network. Via a sparsity-encouraging loss, we relate document-level scores to token-level scores and then we unwind the CNN to produce representative vectors for the tokens. We demonstrate that distances between these vectors can be exploited to establish a mapping between training and test features. We find that distances to nearest true positive, false negative, false positive, and true negative representative vectors from the training set provide a useful and intuitive means of analyzing the data and model. We demonstrate the viability of the approach on a multi-label classification task of electronic health record data, and hypothesize that it will lend itself to a number of additional practical applications in medicine and science.

% % \acks{Many thanks to all collaborators and funders!}

\bibliography{multi_label}

\begin{thebibliography}{17}
\providecommand{\natexlab}[1]{#1}
\providecommand{\url}[1]{\texttt{#1}}
\expandafter\ifx\csname urlstyle\endcsname\relax
  \providecommand{\doi}[1]{doi: #1}\else
  \providecommand{\doi}{doi: \begingroup \urlstyle{rm}\Url}\fi

\bibitem[Belinkov and Glass(2019)]{BelinkovAndGlass-2019-NLP-Analysis-Methods}
Yonatan Belinkov and James Glass.
\newblock {Analysis Methods in Neural Language Processing: A Survey}.
\newblock \emph{Transactions of the Association for Computational Linguistics},
  7:\penalty0 49--72, 2020/04/02 2019.
\newblock \doi{10.1162/tacl_a_00254}.
\newblock URL \url{https://doi.org/10.1162/tacl_a_00254}.

\bibitem[Card et~al.(2019)Card, Zhang, and Smith]{Card-etal-2019-ConformalNLP}
Dallas Card, Michael Zhang, and Noah~A. Smith.
\newblock {Deep Weighted Averaging Classifiers}.
\newblock In \emph{Proceedings of ACM FAT*}, 2019.

\bibitem[Devlin et~al.(2018)Devlin, Chang, Lee, and
  Toutanova]{DevlinEtAl-2018-BERT}
Jacob Devlin, Ming{-}Wei Chang, Kenton Lee, and Kristina Toutanova.
\newblock {{BERT:} Pre-training of Deep Bidirectional Transformers for Language
  Understanding}.
\newblock \emph{CoRR}, abs/1810.04805, 2018.
\newblock URL \url{http://arxiv.org/abs/1810.04805}.

\bibitem[Goldberger et~al.(2000)Goldberger, Amaral, Glass, Hausdorff, Ivanov,
  Mark, Mietus, Moody, Peng, and Stanley]{Physionet-ref-for-MIMIC-3}
Ary~L Goldberger, Luis~AN Amaral, Leon Glass, Jeffrey~M Hausdorff, Plamen~Ch
  Ivanov, Roger~G Mark, Joseph~E Mietus, George~B Moody, Chung-Kang Peng, and
  H~Eugene Stanley.
\newblock {PhysioBank, PhysioToolkit, and PhysioNet: components of a new
  research resource for complex physiologic signals}.
\newblock \emph{Circulation}, 101\penalty0 (23):\penalty0 e215--e220, 2000.

\bibitem[Johnson et~al.(2016)Johnson, Pollard, Shen, Li-wei, Feng, Ghassemi,
  Moody, Szolovits, Celi, and
  Mark]{Johnson-etal-2016-MIMIC-III-main-publication}
Alistair~EW Johnson, Tom~J Pollard, Lu~Shen, H~Lehman Li-wei, Mengling Feng,
  Mohammad Ghassemi, Benjamin Moody, Peter Szolovits, Leo~Anthony Celi, and
  Roger~G Mark.
\newblock {MIMIC-III, a freely accessible critical care database}.
\newblock \emph{Scientific data}, 3:\penalty0 160035, 2016.

\bibitem[Kim(2014)]{Kim-2014-CNN}
Yoon Kim.
\newblock {Convolutional Neural Networks for Sentence Classification}.
\newblock In \emph{Proceedings of the 2014 Conference on Empirical Methods in
  Natural Language Processing (EMNLP)}, pages 1746--1751, Doha, Qatar, October
  2014. Association for Computational Linguistics.
\newblock URL \url{http://www.aclweb.org/anthology/D14-1181}.

\bibitem[Kingma and Ba(2014)]{Kingma-Ba-2014-Adam}
Diederik~P. Kingma and Jimmy Ba.
\newblock {Adam: A Method for Stochastic Optimization}, 2014.

\bibitem[Mikolov et~al.(2013)Mikolov, Sutskever, Chen, Corrado, and
  Dean]{MikolovEtAl-2013-Word2vec}
Tomas Mikolov, Ilya Sutskever, Kai Chen, Greg~S Corrado, and Jeff Dean.
\newblock Distributed representations of words and phrases and their
  compositionality.
\newblock In C.~J.~C. Burges, L.~Bottou, M.~Welling, Z.~Ghahramani, and K.~Q.
  Weinberger, editors, \emph{Advances in Neural Information Processing Systems
  26}, pages 3111--3119. Curran Associates, Inc., 2013.
\newblock URL
  \url{http://papers.nips.cc/paper/5021-distributed-representations-of-words-and-phrases-and-their-compositionality.pdf}.

\bibitem[Mullenbach et~al.(2018)Mullenbach, Wiegreffe, Duke, Sun, and
  Eisenstein]{mullenbach-etal-2018-explainable}
James Mullenbach, Sarah Wiegreffe, Jon Duke, Jimeng Sun, and Jacob Eisenstein.
\newblock {Explainable Prediction of Medical Codes from Clinical Text}.
\newblock In \emph{Proceedings of the 2018 Conference of the North {A}merican
  Chapter of the Association for Computational Linguistics: Human Language
  Technologies, Volume 1 (Long Papers)}, pages 1101--1111, New Orleans,
  Louisiana, June 2018. Association for Computational Linguistics.
\newblock \doi{10.18653/v1/N18-1100}.
\newblock URL \url{https://www.aclweb.org/anthology/N18-1100}.

\bibitem[Pennington et~al.(2014)Pennington, Socher, and
  Manning]{PenningtonEtAl-2014-Glove}
Jeffrey Pennington, Richard Socher, and Christopher Manning.
\newblock {{G}love: Global Vectors for Word Representation}.
\newblock In \emph{Proceedings of the 2014 Conference on Empirical Methods in
  Natural Language Processing ({EMNLP})}, pages 1532--1543, Doha, Qatar,
  October 2014. Association for Computational Linguistics.
\newblock \doi{10.3115/v1/D14-1162}.
\newblock URL \url{https://www.aclweb.org/anthology/D14-1162}.

\bibitem[Pollard(2016)]{Pollard-etal-2016-MIMIC-III-data}
Alistair~EW Pollard, Tom J abd~Johnson.
\newblock {The MIMIC-III Clinical Database}.
\newblock \url{http://dx.doi.org/10.13026/C2XW26}, 2016.

\bibitem[Rei and S{\o}gaard(2018)]{ReiAndSogaard-2018-ZeroShotSeq}
Marek Rei and Anders S{\o}gaard.
\newblock {Zero-Shot Sequence Labeling: Transferring Knowledge from Sentences
  to Tokens}.
\newblock In \emph{Proceedings of the 2018 Conference of the North {A}merican
  Chapter of the Association for Computational Linguistics: Human Language
  Technologies, Volume 1 (Long Papers)}, pages 293--302, New Orleans,
  Louisiana, June 2018. Association for Computational Linguistics.
\newblock \doi{10.18653/v1/N18-1027}.
\newblock URL \url{https://www.aclweb.org/anthology/N18-1027}.

\bibitem[Schmaltz(2019)]{Schmaltz-2019-BLADE-1}
Allen Schmaltz.
\newblock {Detecting Local Insights from Global Labels: Supervised \& Zero-Shot
  Sequence Labeling via a Convolutional Decomposition}.
\newblock arXiv, 2019.
\newblock URL \url{https://arxiv.org/abs/1906.01154}.

\bibitem[Snell et~al.(2017)Snell, Swersky, and
  Zemel]{SnellEtAl-NIPS2017-PrototypicalNets}
Jake Snell, Kevin Swersky, and Richard Zemel.
\newblock {Prototypical Networks for Few-shot Learning}.
\newblock In I.~Guyon, U.~V. Luxburg, S.~Bengio, H.~Wallach, R.~Fergus,
  S.~Vishwanathan, and R.~Garnett, editors, \emph{Advances in Neural
  Information Processing Systems 30}, pages 4077--4087. Curran Associates,
  Inc., 2017.
\newblock URL
  \url{http://papers.nips.cc/paper/6996-prototypical-networks-for-few-shot-learning.pdf}.

\bibitem[Wang et~al.(2018)Wang, Li, Wang, Zhang, Shen, Zhang, Henao, and
  Carin]{Wang-etal-2018-ACL-Label-Embeddings}
Guoyin Wang, Chunyuan Li, Wenlin Wang, Yizhe Zhang, Dinghan Shen, Xinyuan
  Zhang, Ricardo Henao, and Lawrence Carin.
\newblock {Joint Embedding of Words and Labels for Text Classification}.
\newblock In \emph{ACL}, 2018.

\bibitem[Yong et~al.(2010)Yong, Saunders, Olsen, et~al.]{yong2010excess}
Pierre~L Yong, Robert~S Saunders, LeighAnne Olsen, et~al.
\newblock {Excess Administrative Costs}.
\newblock In \emph{The Healthcare Imperative: Lowering Costs and Improving
  Outcomes: Workshop Series Summary}. National Academies Press (US), 2010.

\bibitem[Zeiler(2012)]{Zeiler-2012-Adadelta}
Matthew~D. Zeiler.
\newblock {ADADELTA: An Adaptive Learning Rate Method}.
\newblock \emph{CoRR}, abs/1212.5701, 2012.
\newblock URL \url{http://arxiv.org/abs/1212.5701}.

\end{thebibliography}

\appendix
\section*{Appendix A: MIMIC-III Output Samples}

In Tables~\ref{appendix1-table-MIMIC3-test-exemplar-audit-part1} to~\ref{appendix1-table-MIMIC3-test-exemplar-audit-part3} we illustrate the visualization of the token-level scores and the associated exemplars from the training set with 3 random documents from the test set for the top 50 labels subset. The exemplar is often, but not always a lexical match, and sometimes the surrounding tokens can shed light on the connection between a particular token and its associated exemplar, as with $\boldv_{i^1}$ for label 272.0 in Table~\ref{appendix1-table-MIMIC3-test-exemplar-audit-part1}, where ``high'' (the token of focus) proceeds ``cholesterol'' and the exemplar is associated with ``hypercholesteremia''. 

Often the nearest exemplars are $\boldv_{i^1}$ vectors. In Table~\ref{appendix1-table-MIMIC3-test-exemplar-audit-part2} with Label 39.95 we see a relatively rare example in which the nearest exemplar vector was a $\boldv_{i^4}$ vector. On inspection, we see that $\boldv_{i^1}$ is associated with continuous veno-venous hemofiltration (CVVH),  $\boldv_{i^2}$ is associated with continuous veno-venous hemodialysis (CVVHD), and $\boldv_{i^3}$ with hemodialysis. The next token after the token associated with the $\boldv_{i^4}$ vector is in fact ``cvvhd'', which helps suggest why it was selected. In practice, this would be a case that would be singled out for further review by a human annotator, who would then see that the softmax distribution was relatively diffuse, and make a final decision based on these examples (and the context of the original query).

In Table~\ref{appendix1-table-MIMIC3-test-exemplar-audit-part3} with Label V58.61, we see an example where the query prediction is a false positive and the nearest associated database vector ($\boldv_{i^3}$) is also a false positive. 
 
To a non-specialist, in cases where the model diverges from the ground truth label but the nearest exemplar is a $\boldv_{i^1}$ vector, it is not always clear whether the source of the discrepancy is an idiosyncrasy of ICD-9 coding or noise in the labeling. In some cases, as with Label 96.04 in Table~\ref{appendix1-table-MIMIC3-test-exemplar-audit-part2}, the difference is apparently due to specificity in the choice of a disease or procedure (here, with regard to ``intubation'', using Label 96.71 instead).  

It does seem that closer relative distances (here, a higher value of the ``Normalized Softmax Distance'' included in the tables implies closer similarity) are associated with more reliable coupling between $\boldq_j$ and $\boldv_i$, which is consistent with the empirical results in Table~\ref{table:additional-distance-analysis-all-labels}. Note that these normalized distances are values between 0 and 1, and if the distances between $\boldq_j$ and each of $\boldv_{i^1}$, $\boldv_{i^2}$, $\boldv_{i^3}$, and $\boldv_{i^4}$ were the same, then these distance values would be uniformly 0.25. In practice, if a given prediction was not associated with $\boldv_{i^1}$, or associated with $\boldv_{i^1}$ but with a low relative distance, it could be (tagged in particular to be) shown to a human for further review.

Given that the documents are long, the text is noisy, and the labels are relatively high-dimensional, the output does seem to suggest that such an approach is a useful additional tool for the analysis toolbox.

%%%%%%%%%%%%%%%%%% Exemplar auditing tables 1
\begin{table*} %[b]
\centering
\footnotesize
\begin{tabular}{P{35mm}T{120mm}}
 & Test Document 225\\
\toprule
\textsc{Label 401.9} & {\ttfamily{unspecified essential hypertension}}; Label Frequency in training: 3233\\
\midrule
\textsc{CNN$_{1345}$+mmc} & \ttfamily{...and he went to a osh er where a head ct showed a subdural hematoma patient reports taking two aspirin on the day of admission past medical {\color{blue}history htn[401.9] high cholesterol} social history lawyer lives with...}\\
\textsc{Exemplar $\boldv_{i^1}$ [401.9]}  & Normalized Softmax Distance: 0.604\\
\textsc{Exemplar $\boldv_{i^1}$ [401.9], Train Doc. 2201}  & \ttfamily{...but this was not covered by insurance and hence he does not take it past medical history {\color{blue}htn[401.9]} chol bph right renal cyst social history he is...}\\
\midrule
\textsc{Label 272.0} & {\ttfamily{pure hypercholesterolemia}}; Label Frequency in training: 926\\
\midrule
\textsc{CNN$_{1345}$+mmc} & \ttfamily{...past medical history htn {\color{blue}high[272.0]} cholesterol social history lawyer lives with...}\\
\textsc{Exemplar $\boldv_{i^1}$ [272.0]}  & Normalized Softmax Distance: 0.384\\
\textsc{Exemplar $\boldv_{i^1}$ [272.0], Train Doc. 6565}  & \ttfamily{...and the hematuria has since resolved he has been experiencing insomnia for the past month past medical history htn {\color{blue}hypercholesteremia[272.0] } etoh daily use gout...}\\
\bottomrule
\end{tabular}
\caption{\small{Exemplar auditing output for the first of three random documents from the test set for the top 50 labels subset for the \textsc{CNN$_{1345}$+mmc} model. In this case, both ground truth labels are correctly predicted and all of the exemplar vectors from training correspond to $\boldv_{i^1}$ (and the remaining vectors are not displayed). These are short snippets of longer documents. We further truncate subsequent instances of the same document (token scores are label specific per document). We color highlights associated with correct predictions at the document-level in blue, and those associated with incorrect predictions in red, but note that ground-truth token-level labels are not available here. Labels, associated descriptions, and label frequencies in training are provided. The tokens associated with $\boldq_j$ and $\boldv_i$ are marked with brackets (with the ICD-9 code), and the identity (TP, FN, FP, etc.) of $\boldv_{i^*}$ is specified along with the normalized softmax distance (where greater values imply closer similarity).}}
\label{appendix1-table-MIMIC3-test-exemplar-audit-part1}
\end{table*}
%%%%%%%%%%%%%%%%%% Exemplar auditing tables

%%%%%%%%%%%%%%%%%% Exemplar auditing tables 2
\begin{table*} %[b]
\centering
\scriptsize
\begin{tabular}{P{35mm}T{120mm}}
 & Test Document 314\\
\toprule
\textsc{Label 96.04} & {\ttfamily{insertion of endotracheal tube}}; Label Frequency in training: 1581\\
\midrule
\textsc{CNN$_{1345}$+mmc} & \ttfamily{...chief complaint depakote overdose major surgical or invasive procedure {\color{red}intubation[96.04]} hemodialysis femoral and jugular central line placements history of present illness the patient is a year old female with a reported history of alcohol abuse and bipolar disorder who...}\\
\textsc{Exemplar $\boldv_{i^1}$ [96.04]}  & Normalized Softmax Distance: 0.328\\
\textsc{Exemplar $\boldv_{i^1}$ [96.04], Train Doc. 2981}  & \ttfamily{...chief complaint attempted suicide major surgical or invasive procedure {\color{blue}intubation[96.04]} wrist laceration repair history of present illness year old man presented to the hospital1 ed in the setting of a reported suicide attempt via laceration to his right wrist...}\\
\midrule
\textsc{Label 96.71} & {\ttfamily{continuous invasive mechanical ventilation for less than 96 consecutive hours}}; Label Frequency in training: 1395\\
\midrule
\textsc{CNN$_{1345}$+mmc} & \ttfamily{...she developed progressive confusion to the point of somlanence she was reported to vomit she was subsequently {\color{blue}intubated[96.71]} for airway protection and transfered to the hospital1 ed for further manegment...}\\
\textsc{Exemplar $\boldv_{i^1}$ [96.71]}  & Normalized Softmax Distance: 0.391\\
\textsc{Exemplar $\boldv_{i^1}$ [96.71], Train Doc. 1338}  & \ttfamily{...this is a yo f s p suicide attempt with cymbalta klonopin alcohol and cyproheptadine now s p extubation and medically stable she was initially {\color{blue}intubated} known firstname ed for somnolence she was extubated on without further events...}\\
\midrule
\textsc{Label 39.95} & {\ttfamily{hemodialysis}}; Label Frequency in training: 549\\
\midrule
\textsc{CNN$_{1345}$+mmc} & \ttfamily{...chief complaint depakote overdose major surgical or invasive procedure {\color{blue}intubation hemodialysis[39.95]} femoral and jugular central line placements history of present illness the patient is a year old female with a reported history of alcohol abuse and bipolar disorder who...}\\
%v1
\textsc{Exemplar $\boldv_{i^1}$ [39.95]}  & Normalized Softmax Distance: 0.263\\
\textsc{Exemplar $\boldv_{i^1}$ [39.95], Train Doc. 1379}  & \ttfamily{...invasive procedure leukophareisis {\color{blue}cvvh[39.95]} history of present illness...}\\
%v2
\textsc{Exemplar $\boldv_{i^2}$ [39.95] }  & Normalized Softmax Distance: 0.150\\
\textsc{Exemplar $\boldv_{i^2}$ [39.95], Train Doc. 5439}  & \ttfamily{...was not resumed gu uop augmented with **cvvhd**[39.95] perioperatively from date range creatinine stablilized...}\\
%v3
\textsc{Exemplar $\boldv_{i^3}$ [39.95] }  & Normalized Softmax Distance: 0.206\\
\textsc{Exemplar $\boldv_{i^3}$ [39.95], Train Doc. 846}  & \ttfamily{...the patient was noticed to have decreased mental status after **hemodialysis**[39.95] yesterday which worsened on the day of presentation...}\\
%v4
\textsc{Exemplar $\boldv_{i^4}$ [39.95] }  & Normalized Softmax Distance: 0.381\\
\textsc{Exemplar $\boldv_{i^4}$ [39.95], Train Doc. 517}  & \ttfamily{...major surgical or invasive procedure intubation r {\color{red}ij[39.95]} cvvhd history of present illness...}\\
\midrule
\textsc{Label 311} & {\ttfamily{depressive disorder, not elsewhere classified}}; Label Frequency in training: 493\\
\midrule
\textsc{CNN$_{1345}$+mmc} & \ttfamily{...and folic acid for vitamin supplementation {\color{red}depression[311]} she readily admitted to her overdose being an attempt at suicide and expressed considerable remorse in this action...}\\
\textsc{Exemplar $\boldv_{i^1}$ [311]}  & Normalized Softmax Distance: 0.374\\
\textsc{Exemplar $\boldv_{i^1}$ [311], Train Doc. 2981}  & \ttfamily{...year old man with polysubstance dependence on suboxone and {\color{blue}depression[311]} admitted after suicide attempt...}\\
\bottomrule
\end{tabular}
\caption{\footnotesize{Exemplar auditing output for the second of three random documents from the test set for the top 50 labels subset for the \textsc{CNN$_{1345}$+mmc} model, with similar formatting as Table~\ref{appendix1-table-MIMIC3-test-exemplar-audit-part1}. In this case, of the 5 ground-truth labels (96.71,285.9,276.2,39.95,305.1), two were correctly predicted, and the remainder were false negatives (not shown). Additionally, two predictions were false positive. For Label 311, we display all 4 exemplar vectors, as this is a relatively rare case in which $\boldv_{i^4}$ was selected. Symbols ** are used for exemplars associated with $\boldv_{i^2}$ or $\boldv_{i^3}$.}}
\label{appendix1-table-MIMIC3-test-exemplar-audit-part2}
\end{table*}
%%%%%%%%%%%%%%%%%% Exemplar auditing tables

%%%%%%%%%%%%%%%%%% Exemplar auditing tables 3
\begin{table*} %[b]
\centering
\footnotesize
\begin{tabular}{P{35mm}T{120mm}}
 & Test Document 316\\
\toprule
\textsc{Label 401.9} & {\ttfamily{unspecified essential hypertension}}; Label Frequency in training: 3233\\
\midrule
\textsc{CNN$_{1345}$+mmc} & \ttfamily{...has never been diagnosed with dementia past medical history pmh atrial fibrillation on coumadin {\color{blue}htn[401.9]} acoustic neuroma resected years ago on the right hyperthyroidism now hypothyroid after iodine therapy...}\\
\textsc{Exemplar $\boldv_{i^1}$ [401.9]}  & Normalized Softmax Distance: 0.420\\
\textsc{Exemplar $\boldv_{i^1}$ [401.9], Train Doc. 951}  & \ttfamily{...collar was removed at outside facility past medical history iddm a fib on coumadin {\color{blue}htn[401.9]} mild aortic stenosis cva in past history of old small reportedly lacunar infarcts etoh abuse...}\\
\midrule
\textsc{Label 427.31} & {\ttfamily{atrial fibrillation}}; Label Frequency in training: 1992\\
\midrule
\textsc{CNN$_{1345}$+mmc} & \ttfamily{...dementia past medical history {\color{blue}pmh atrial fibrillation[427.31] on coumadin} htn acoustic...}\\
\textsc{Exemplar $\boldv_{i^1}$ [427.31]}  & Normalized Softmax Distance: 0.540\\
\textsc{Exemplar $\boldv_{i^1}$ [427.31], Train Doc. 2064}  & \ttfamily{...pt is a age over yo female with atrial {\color{blue}fibrillation[427.31]}  on coumadin htn and csf who fell at her nursing home...}\\
\midrule
\textsc{Label 599.0} & {\ttfamily{urinary tract infection, site not specified}}; Label Frequency in training: 1067\\
\midrule
\textsc{CNN$_{1345}$+mmc} & \ttfamily{...she was treated with cipro for days for an e {\color{blue}coli uti[599.0]}  on hd she removed her foley catheter and there was no evidence of trauma...}\\
\textsc{Exemplar $\boldv_{i^1}$ [599.0]}  & Normalized Softmax Distance: 0.304\\
\textsc{Exemplar $\boldv_{i^1}$ [599.0], Train Doc. 5784}  & \ttfamily{...the patient s primary oncologist was notified of her admission {\color{blue}uti[599.0]} pt had postive urine culture after leukocytosis...}\\
... \\
\textsc{Exemplar $\boldv_{i^3}$ [599.0]}  & Normalized Softmax Distance: 0.367\\
\textsc{Exemplar $\boldv_{i^3}$ [599.0], Train Doc. 4599}  & \ttfamily{...wenckebach rhythm delirium and an e coli **uti**[599.0] treated with levofloxacin and then bactrim with the concern...}\\
\midrule
\textsc{Label 244.9} & {\ttfamily{unspecified acquired hypothyroidism}}; Label Frequency in training: 761\\
\midrule
\textsc{CNN$_{1345}$+mmc} & \ttfamily{...resected years ago on the right hyperthyroidism now {\color{red}hypothyroid[244.9]} after iodine therapy years ago macular degeneration left sided hearing loss...}\\
\textsc{Exemplar $\boldv_{i^1}$ [244.9]}  & Normalized Softmax Distance: 0.308\\
\textsc{Exemplar $\boldv_{i^1}$ [244.9], Train Doc. 5784}  & \ttfamily{...ulcer disease colonic adenoma goiter with {\color{blue}hypothyroidism[244.9]} osteoporosis osteoarthritis...}\\
\midrule
\textsc{Label V58.61} & {\ttfamily{long-term (current) use of anticoagulants}}; Label Frequency in training: 604\\
\midrule
\textsc{CNN$_{1345}$+mmc} & \ttfamily{...atrial fibrillation {\color{red}on[V58.61]} coumadin htn...}\\
\textsc{Exemplar $\boldv_{i^1}$ [V58.61]}  & Normalized Softmax Distance: 0.327\\
\textsc{Exemplar $\boldv_{i^1}$ [V58.61], Train Doc. 5702}  & \ttfamily{...atrial fibrillation {\color{blue}on[V58.61]} coumadin osteoarthritis s p hemithyroidectomy...}\\
... \\
\textsc{Exemplar $\boldv_{i^3}$ [V58.61]}  & Normalized Softmax Distance: 0.394\\
\textsc{Exemplar $\boldv_{i^3}$ [V58.61], Train Doc. 951}  & \ttfamily{...iddm a fib **on**[V58.61] coumadin htn mild aortic stenosis cva...}\\
\bottomrule
\end{tabular}
\caption{\footnotesize{Exemplar auditing output for the third of three random documents from the test set for the top 50 labels subset for the \textsc{CNN$_{1345}$+mmc} model, with similar formatting as Tables~\ref{appendix1-table-MIMIC3-test-exemplar-audit-part1} and ~\ref{appendix1-table-MIMIC3-test-exemplar-audit-part2}. In this case, of the 3 ground-truth labels (401.9,427.31,599.0), all three were correctly predicted, but there were also two false positives. For Labels 599.0 and V58.61, the nearest exemplars were $\boldv_{i^3}$ vectors.}}
\label{appendix1-table-MIMIC3-test-exemplar-audit-part3}
\end{table*}
%%%%%%%%%%%%%%%%%% Exemplar auditing tables

\end{document}